\documentclass[10pt,twocolumn,letterpaper]{article}

\usepackage[pagenumbers]{cvpr} 

\usepackage{graphicx}
\usepackage{amsmath}
\usepackage{amssymb}
\usepackage{booktabs}
\usepackage[dvipsnames]{xcolor}
\usepackage[pagebackref,breaklinks,colorlinks,citecolor=citecolor]{hyperref}
\definecolor{citecolor}{HTML}{62BDB6}

\usepackage[capitalize]{cleveref}
\crefname{section}{Sec.}{Secs.}
\Crefname{section}{Section}{Sections}
\Crefname{table}{Table}{Tables}
\crefname{table}{Tab.}{Tabs.}

\def\cvprPaperID{***} 
\def\confName{CVPR}
\def\confYear{2022}

\newcommand{\gf}[1]{{\textbf{\color{Green}{#1}}}} 
\newcommand{\bd}[1]{{\color{Blue}{\underline{#1}}}} 

\begin{document}

\title{Investigating Tradeoffs in Real-World Video Super-Resolution}
\author{Kelvin C.K. Chan\qquad Shangchen Zhou\qquad  Xiangyu Xu\qquad Chen Change Loy\\
S-Lab, Nanyang Technological University\\
{\tt\small \{chan0899, s200094, xiangyu.xu, ccloy\}@ntu.edu.sg}
}

\thispagestyle{empty}
\twocolumn[{%
      \renewcommand\twocolumn[1][]{#1}%
      \vspace{-1cm}
      \maketitle
      \vspace{-1cm}
      \begin{center}
        \centering
        \includegraphics[width=0.98\textwidth]{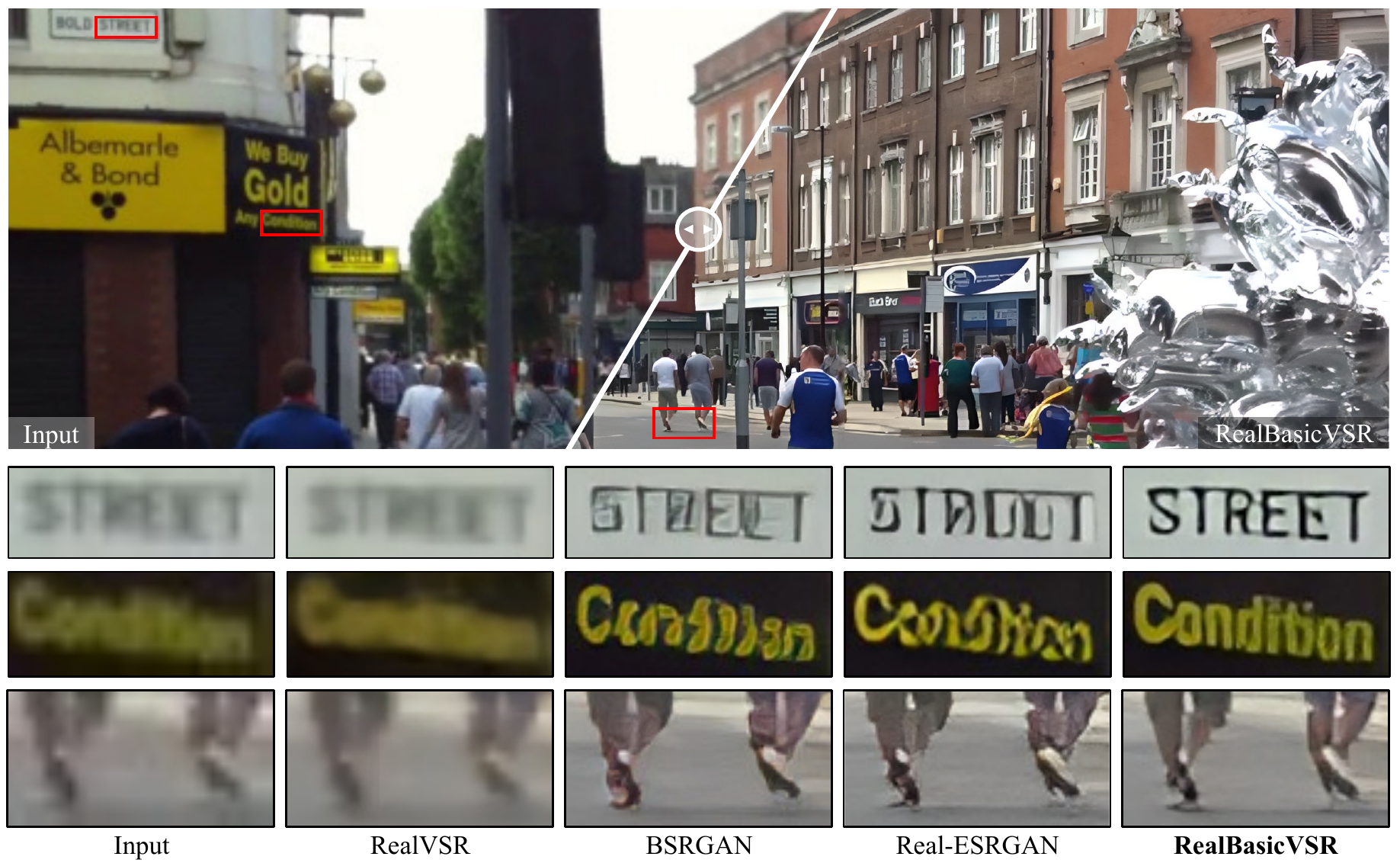}\vspace{0.1cm}
        \vskip -0.2cm
        \captionof{figure}{\textbf{Results on a Real-World Video.} In this work, we investigate various tradeoffs caused by the complex and diverse degradations in real-world VSR. Such tradeoffs are largely neglected in the literature. We propose simple yet effective solutions to the tradeoffs, and the resulting model \textit{RealBasicVSR} acts as a strong baseline for real-world VSR. \textbf{(Zoom-in for best view)}
        }
        \label{fig:teaser}
      \end{center}
    }]

\begin{abstract}
  \vspace{-0.5cm}
  The diversity and complexity of degradations in real-world video super-resolution (VSR) pose non-trivial challenges in inference and training.
  First, while long-term propagation leads to improved performance in cases of mild degradations, severe in-the-wild degradations could be exaggerated through propagation, impairing output quality. To balance the tradeoff between detail synthesis and artifact suppression, we found an image pre-cleaning stage indispensable to reduce noises and artifacts prior to propagation. Equipped with a carefully designed cleaning module, our RealBasicVSR outperforms existing methods in both quality and efficiency (Fig.~\ref{fig:teaser}).
  Second, real-world VSR models are often trained with diverse degradations to improve generalizability, requiring increased batch size to produce a stable gradient. Inevitably, the increased computational burden results in various problems, including 1) speed-performance tradeoff and 2) batch-length tradeoff. To alleviate the first tradeoff, we propose a stochastic degradation scheme that reduces up to 40\% of training time without sacrificing performance. We then analyze different training settings and suggest that employing longer sequences rather than larger batches during training allows more effective uses of temporal information, leading to more stable performance during inference.
  To facilitate fair comparisons, we propose the new VideoLQ dataset, which contains a large variety of real-world low-quality video sequences containing rich textures and patterns. Our dataset can serve as a common ground for benchmarking. Code, models, and the dataset will be made publicly available at https://github.com/ckkelvinchan/RealBasicVSR.
\end{abstract}


\section{Introduction}
In real-world video super-resolution (VSR), we aim at increasing the resolution of videos containing unknown degradations. The diversity of degradations in this task poses significant challenges in designing benchmarks and training settings, and hence earlier works assume either synthetic~\cite{chan2021basicvsr,chan2021basicvsrpp,wang2019edvr} or camera-specific~\cite{yang2021real} degradations and focus on network designs. Although these works achieve remarkable success in restricted settings, the designs for these over-simplified scenarios cannot generalize well to the complex degradations in the wild.
In addition, the complexity and diversity of degradations in real-world VSR introduce extra obstacles in both inference and training, including artifact amplification and increased computational budgets. This paper dives into the problems and tradeoffs in real-world VSR to share useful experiences in addressing the task.

It is shown by Chan \etal~\cite{chan2021basicvsr} that
long-term information is beneficial to restoration. However, in real-world VSR, such information could also result in exaggerated artifacts, owing to error accumulation during propagation.
This phenomenon leads to a tradeoff between \textit{enhancing details} and \textit{suppressing artifacts}, since the synthesizing power of a network comes at the cost of amplifying noises and artifacts.
In this work, we show that a simple solution can sufficiently remedy this tradeoff. In particular, we place an \textit{image cleaning} module prior to propagation for removing degradations in the input images. The resulting model \textit{RealBasicVSR} avoids amplification of artifacts and achieves improved output quality while maintaining simplicity.
We further develop a dynamic refinement scheme that repeatedly applies the cleaning module to remove excessive degradations in the inputs. Our scheme allows a flexible tradeoff between \textit{smoothness} and \textit{detailedness}, which can be adjusted based on a pre-defined threshold or user preference. A systematic analysis of different combinations of losses and architectures is conducted to demonstrate the significance of our designs.

\begin{figure*}[t]
    \begin{center}
        \includegraphics[width=0.95\textwidth]{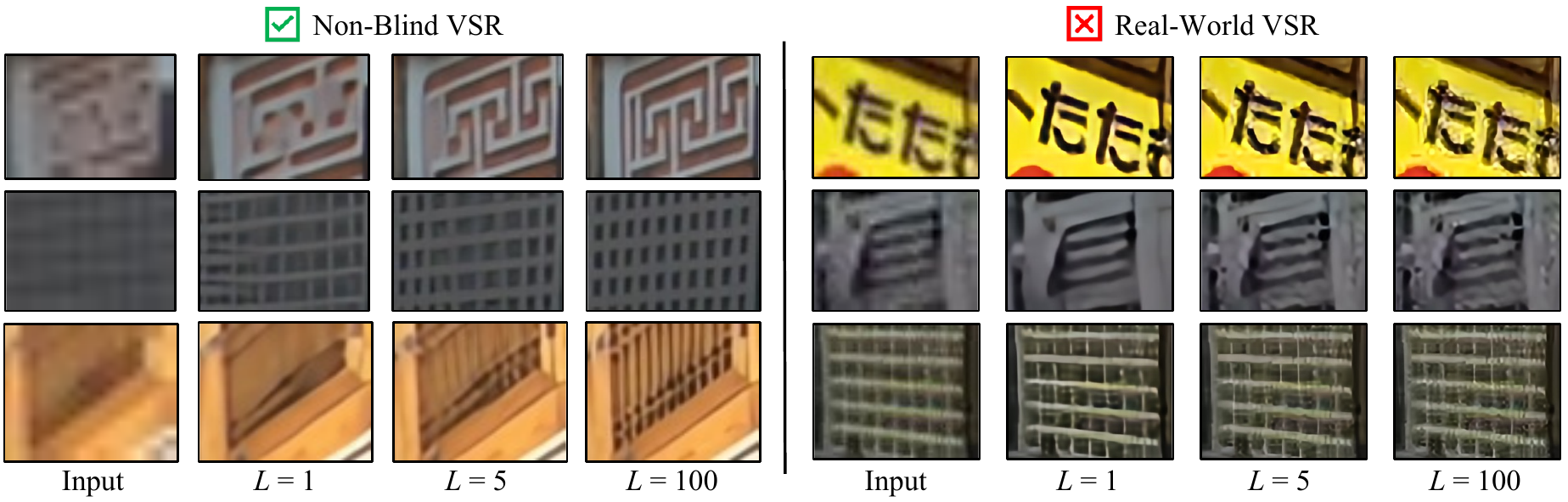}
        \vspace{-0.3cm}
        \caption{\textbf{Effects of Long-Term Propagation.} While employing long-term information leads to improved performance in non-blind VSR, propagation in real-world scenarios could lead to undesirable artifacts. $L$ denotes the sequence length. \textbf{(Zoom-in for best view)}}
        \label{fig:motivation}
    \end{center}
    \vspace{-0.5cm}
\end{figure*}

Real-world VSR models are generally trained with diverse degradations to improve generalizability, and hence they are often trained with increased batch size to ensure stable gradient. As a result, real-world VSR usually requires a longer training time and more immense computational resources than the non-blind counterpart. This work inspects two tradeoffs in real-world VSR to improve training efficiency, hence shortening research cycles.

First, with increased batch size, training with long sequences is prohibitive owing to the I/O bottleneck induced by hardware limitations. The bottleneck is often alleviated by reducing either the batch size or sequence length, which results in degraded performance. To ameliorate the problem, we propose a \textit{stochastic degradation} scheme that effectively reduces the I/O bottleneck without sacrificing the output quality. Notably, our degradation scheme yields up to \emph{40\%} reduction of training time in comparison to the conventional training scheme.

Second, with a fixed computational budget, the increased batch size in real-world VSR inevitably decreases sequence length during training. We are interested in the tradeoff between them with an aim to search for a more effective setting. To this end, we compare models trained with different combinations of batch sizes and sequence lengths. We conclude that networks trained with longer sequences rather than larger batches could more effectively employ the long-term information in the input sequence, improving stability.

In addition to the studies above, we introduce a new benchmark for real-world VSR. Most existing benchmarks~\cite{liu2014bayesian,nah2019ntire,xue2019video,yi2019progressive} are constructed by contaminating the high-resolution (HR) videos with pre-defined degradations.
The most recent RealVSR dataset~\cite{yang2021real} exploits the dual-camera system in iPhone to capture paired data. Yet, the RealVSR dataset consists of only degradations specifically for the iPhone camera.
With only pre-defined degradations, the benchmarks mentioned above cannot accurately reflect the generalizability of the models on real-world videos.
In this work, we propose \textit{\mbox{VideoLQ}}, a real-world video dataset consisting of diverse LR videos to cover various contents, resolutions, and degradations.
Our dataset could serve as a common benchmark for future methods. We test existing methods on our datasets. Their quantitative and qualitative results and our dataset will be released for ease of future research.
\section{Related Work}
\noindent\textbf{Video Super-Resolution.}
Most existing VSR methods~\cite{cao2021video,chan2021basicvsr,chan2021understanding,chan2021basicvsrpp,isobe2020video1,isobe2020video,isobe2020revisiting,jo2018deep,wang2019edvr,xue2019video,yang2021real,xu2019towards} are trained with pre-defined degradations (\eg, either synthetic~\cite{liu2014bayesian,nah2019ntire,xue2019video,yi2019progressive} or camera-specific~\cite{yang2021real}),
and they deteriorate significantly when handling unknown degradations in reality.
However, extending from non-blind VSR to real-world VSR is non-trivial due to various problems induced by the complex degradations in the wild. For example, artifact amplification during long-term propagation limits the performance of existing VSR methods, and increased computational costs lengthen research cycles. In this work, we investigate the challenges in both inference and training, and provide respective solutions to the challenges.

\noindent\textbf{Real-World Super-Resolution.}
Extended from synthetic settings~\cite{chan2021glean,dong2014learning,dong2016image,dong2016accelerating,wang2018recovering,zhou2020cross}, \textit{blind} super-resolution~\cite{gu2019blind,hui2021learning,ji2020real,liang2021mutual,liang2021flow,luo2020unfolding,xu2019towards} assumes the inputs are degraded by a known process with unknown parameters. The networks are trained with a pre-defined set of degradations with the parameters chosen at random. While the trained networks are able to restore images/videos with a range of degradations, the variation of degradations is often limited, and the generalizability to real-world degradations is in doubt.
Two recent studies~\cite{wang2021real,zhang2021designing} propose to employ more diverse degradations for data augmentation during training. By using ESRGAN~\cite{wang2018esrgan} with no changes in architecture, these two methods demonstrate promising performance in real-world images.
However, we find that such a direct extension at the data augmentation level is not feasible in real-world VSR as the network tends to amplify the noise and artifacts. In this work, we investigate the cause and propose a simple yet effective \textit{image cleaning} module to remedy the problem. Equipped with the cleaning module, \textit{RealBasicVSR} outperforms existing works, including~\cite{wang2021real,zhang2021designing}, in both quality and efficiency.

\noindent\textbf{Input Pre-Processing.}
In this study, we find that a seemingly trivial image cleaning module is critical to remove degradations prior to propagation and suppress artifacts in the outputs. The merit is even more profound in the existence of long-term propagation.
In SISR, similar notions~\cite{kim2020unsupervised,lugmayr2019unsupervised,maeda2020unpaired,rad2021benefiting,yuan2018unsupervised} have been discussed for unsupervised settings.
Despite the success in the unsupervised paradigm, input pre-processing in supervised settings and in VSR are not explored. In contrast to the works above, we focus on an entirely different supervised VSR setting to remove degradations that are amplified during long-term propagation. In addition, we devise a dynamic refinement scheme, which has not been explored in previous works, to remove excessive degradations by repeatedly applying the cleaning module during inference.
We also conduct systematic analysis on our image cleaning module and refinement scheme to verify its effectiveness and provide insights for future studies.

\section{Tradeoff in Inference}
\subsection{Motivation}
VSR networks boost details and improve perceptual quality through aggregating information from multiple frames. But in the case of unseen degradations, the network may fail to distinguish unwanted artifacts from favorable details. Therefore, such artifacts and noises are enhanced through temporal propagation.
To verify our hypothesis, we retrain BasicVSR~\cite{chan2021basicvsr} for real-world VSR. BasicVSR accepts sequences with arbitrary lengths, allowing us to explore the effects of temporal propagation by adjusting the sequence length.
We train BasicVSR with the degradation scheme and settings of Real-ESRGAN~\cite{wang2021real}, which are shown effective in real-world SISR.

As shown in Fig.~\ref{fig:motivation}~(left), in non-blind settings, when the sequence length $L$ increases, BasicVSR is able to aggregate useful information through long-term propagation, generating more details in the outputs.
In contrast, in real-world VSR, while propagation helps enhance details in cases of mild degradations, it is observed in Fig.~\ref{fig:motivation}~(right) that propagating through a longer sequence could amplify noises and artifacts. For instance, when restoring the sequence using only one frame, BasicVSR is able to remove the noises in the inputs and produce smooth outputs, but propagating across the entire sequence leads to outputs with severe artifacts.

In real-world VSR, temporal propagation is a double-edged sword. While employing long-term information helps synthesize fine details, it can also introduce unpleasant artifacts. Clearly, there is a tradeoff between \textit{enhancing details} and \textit{suppressing artifacts}.

\begin{figure}[t]
    \begin{center}
        \includegraphics[width=0.49\textwidth]{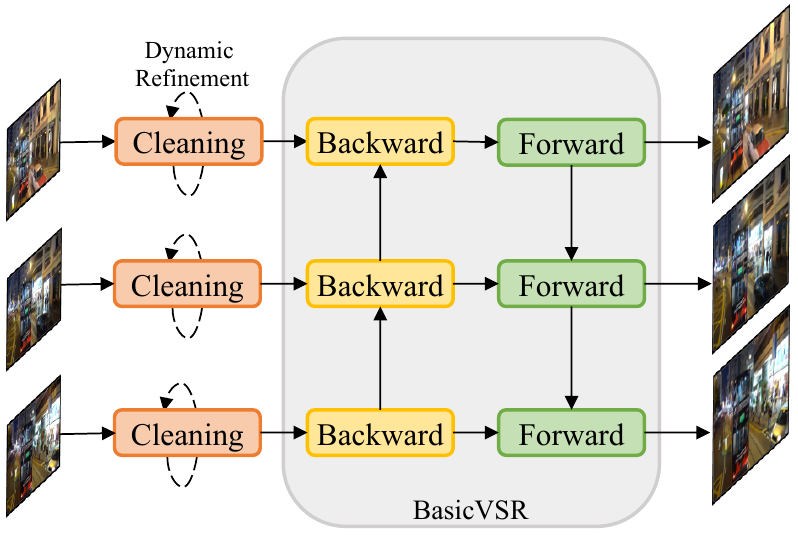}
        \caption{\textbf{Overview of RealBasicVSR.} The input images are first independently passed to our image cleaning module. The clean sequence is then passed to the VSR network (\ie, BasicVSR~\cite{chan2021basicvsr}). Note that the whole network is trained end-to-end.}
        \label{fig:overview}
    \end{center}
    \vspace{-0.4cm}
\end{figure}

\subsection{Input Pre-Cleaning for Real-World VSR}
Motivated by the above, we propose a simple plugin to suppress degradations prior to temporal propagation. The high-level idea is to ``clean'' the input sequence so that the degradations in the inputs have a weaker effect on the subsequent VSR network. Despite being conceptually simple, the designs of the module require special care. More analysis of our cleaning module can be found in Sec.~\ref{subsec:analysis}.

\vspace{0.05cm}
\noindent\textbf{Formulation.}
An overview is shown in Fig.~\ref{fig:overview}.
The image cleaning module is used prior to BasicVSR~\cite{chan2021basicvsr}. The input images are first independently passed to the cleaning module for degradation removal. Let $x_i$ be the $i$-th image of the input sequence, and $C$ be our image cleaning module, we have
\begin{equation}
    \tilde{x}_i = C\left(x_i\right).
\end{equation}
The clean sequence is then passed to the VSR network $S$ for super-resolution:
\begin{equation}
    \{y_i\} = S\left(\{\tilde{x}_i\}\right).
\end{equation}
We adopt BasicVSR~\cite{chan2021basicvsr} in this work because of its promising performance in non-blind VSR through long-term propagation, and its simplicity in architecture.

To guide the image cleaning module, we constrain the outputs of the cleaning module with a low-resolution ground-truth:
\begin{equation}
    \label{eq:cleaning_loss}
    \mathcal{L}_{clean} = \sum_i\rho\left(\tilde{x}_i - d(z_i)\right),
\end{equation}
where $z_i$ is the ground-truth high-resolution image, and $d$ is a downsampling operator\footnote{The \texttt{area} mode in PyTorch.}. Here $\rho$ represents the Charbonnier loss~\cite{charbonnier1994two}. In addition to the cleaning loss, we also use the output fidelity loss to guide the cleaning module.
\begin{equation}
    \label{eq:hr_loss}
    \mathcal{L}_{out} = \sum_i\rho\left(y_i - z_i\right).
\end{equation}
Note that the cleaning module is detached from the perceptual loss~\cite{johnson2016perceptual} and adversarial loss~\cite{goodfellow2014generative} when we finetune the network with these two losses.

\noindent\textbf{Dynamic Refinement.}
A single pass of input to the cleaning module cannot effectively remove the excessive degradations in many challenging cases. A simple yet effective method is to suppress the degradations further with another pass to the cleaning module.
Formally, we design a refinement scheme that dynamically removes the degradations in test time:
\begin{equation}
    \label{eq:dynamic}
    \begin{cases}
        \tilde{x}^{j+1}_i = C(\tilde{x}^{j}_i)\quad & \text{if } mean\left(|\tilde{x}^{j}_i - \tilde{x}^{j-1}_i|\right) \geq \theta, \\
        \tilde{x}_i = \tilde{x}^{j}_i               & \text{otherwise},
    \end{cases}
\end{equation}
where $\tilde{x}_i^0=x_i$, and $\theta$ is a pre-determined stopping criteria. We find that $\theta{=}1.5$ for non GAN-based models and $\theta{=}5$ for GAN-based models
are reasonable settings.

\noindent\textbf{Architecture.}
In this work, we simply use a stack of residual blocks~\cite{he2016deep} as the cleaning module. It is noteworthy that while our cleaning module is conceptually straightforward, it cannot take arbitrary designs, as we will discuss in Sec.~\ref{subsec:analysis}.
In our design, the role of artifact suppression of VSR network is shared by the cleaning module, and hence a lighter VSR network can be adopted. In our experiments, we reduce the residual blocks in BasicVSR from $60$ to $40$ to maintain a comparable complexity.

\begin{figure}[t]
    \begin{center}
        \includegraphics[width=0.48\textwidth]{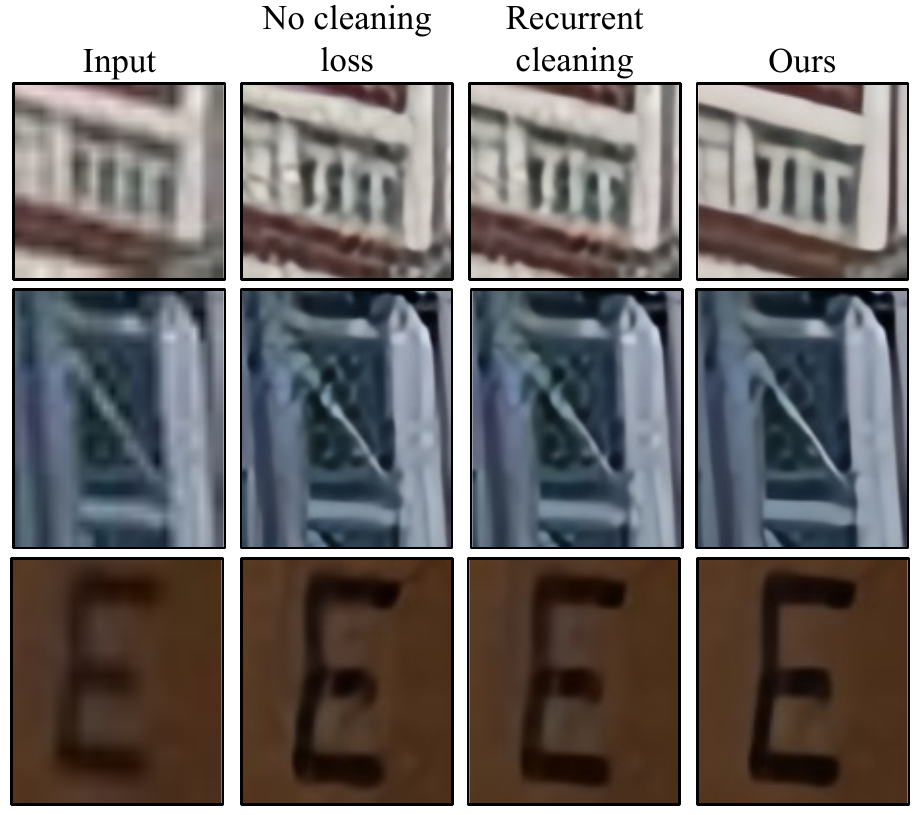}
        \caption{\textbf{Analysis of the Cleaning Module.} The proposed cleaning loss plays an important role in removing the artifacts. The design of the cleaning module requires special care. An alternative model that uses a recurrent structure fails to remove the artifacts. \textbf{(Zoom-in for best view)}}
        \label{fig:analysis}
    \end{center}
    \vspace{-0.4cm}
\end{figure}

\subsection{Analysis of Input Pre-Cleaning}
\label{subsec:analysis}
\noindent\textbf{Designs.}
We study the effects of the proposed image cleaning loss and the architecture of the cleaning module. Examples are shown in Fig.~\ref{fig:analysis}.

First, we train RealBasicVSR with the image cleaning loss (Eqn.~\eqref{eq:cleaning_loss}) removed. When the loss is removed, RealBasicVSR can be regarded as a single-stage network as BasicVSR. The network exaggerates the noises and artifacts, and the original content is distorted, showing the importance of the image cleaning loss. Note that additional losses such as adversarial loss and perceptual loss can be adopted, but we find the simplest pixelwise loss suffices.

Second, we keep the image cleaning loss and replace our cleaning module with a recurrent network. Even with the cleaning loss, the network fails to remove the unwanted degradations, also leading to distorted outputs. This observation is coherent to our hypothesis that video-based networks tend to exaggerate artifacts through aggregation, and demonstrates the importance of adopting an image-based network as the cleaning module.
When compared to the aforementioned variants, our designs produce much smoother outputs, and preserve more image content.

\begin{figure}[t]
    \begin{center}
        \includegraphics[width=0.45\textwidth]{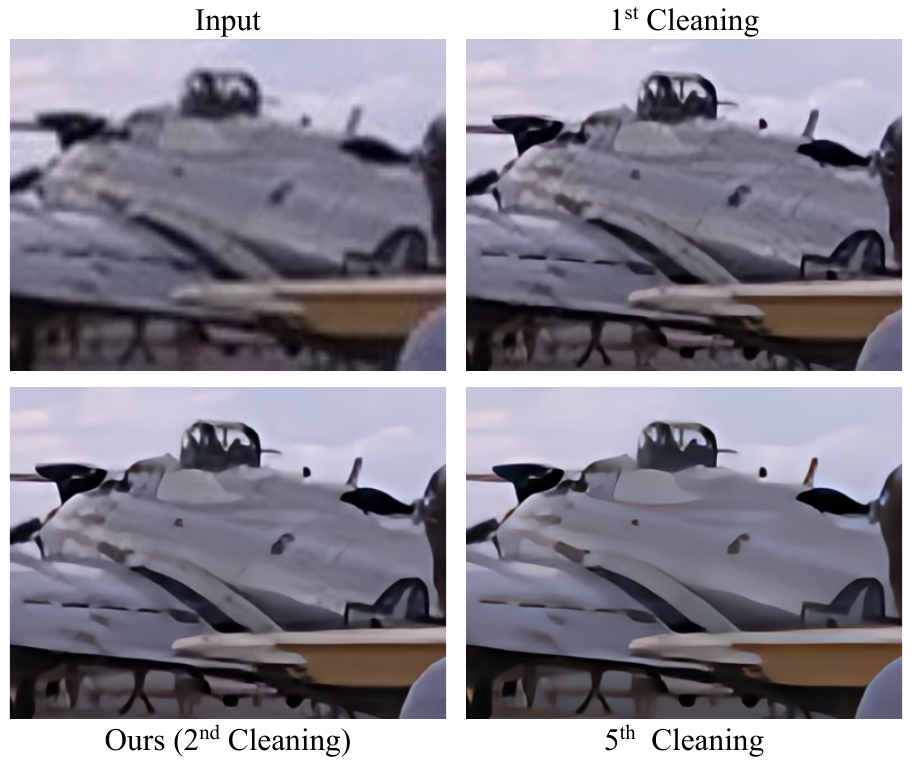}
        \caption{\textbf{Effect of Dynamic Refinement.} Our dynamic refinement scheme automatically stops the cleaning process to avoid over-smoothing and unnaturally flat regions. More examples are provided in the supplementary material. \textbf{(Zoom-in for best view)}}
        \label{fig:cleaning}
    \end{center}
    \vspace{-0.6cm}
\end{figure}
\begin{figure}[t]
    \begin{center}
        \includegraphics[width=0.48\textwidth]{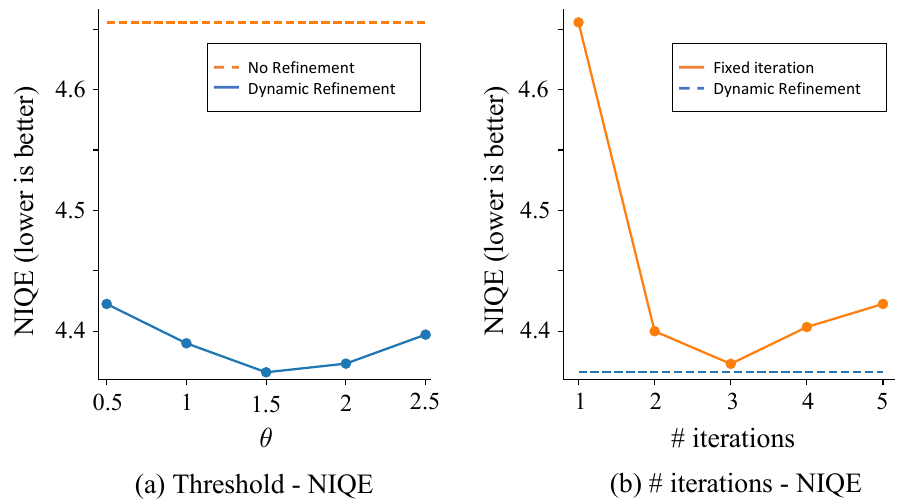}
        \vspace{-0.3cm}
        \caption{\textbf{Ablations on Refinement.}
            (a) The NIQE is significantly lower with our dynamic refinement. The thresholds control the levels of details, leading to different NIQE. (b) Our dynamic refinement scheme obtains a better NIQE than fixed iterations. NIQE is computed on our VideoLQ dataset.}
        \label{fig:cleaning_plot}
    \end{center}
    \vspace{-0.6cm}
\end{figure}
\begin{figure}[t]
    \begin{center}
        \includegraphics[width=0.49\textwidth]{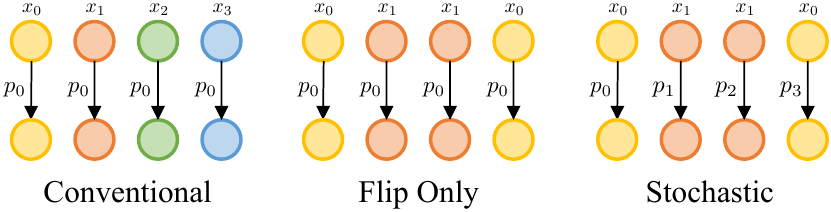}
        \caption{\textbf{Stochastic Degradation Scheme.} By loading fewer frames per iteration and using temporally-varying degradations, our stochastic degradation scheme reduces the training time by \emph{40\%} without sacrificing performance. Each circle represents one video frame, and $p_i=p_{i-1} + r_i$ (Eqn.~\eqref{eq:stoc}). }
        \label{fig:stochastic}
    \end{center}
    \vspace{-0.7cm}
\end{figure}
\begin{figure}[t]
    \begin{center}
        \includegraphics[width=0.45\textwidth]{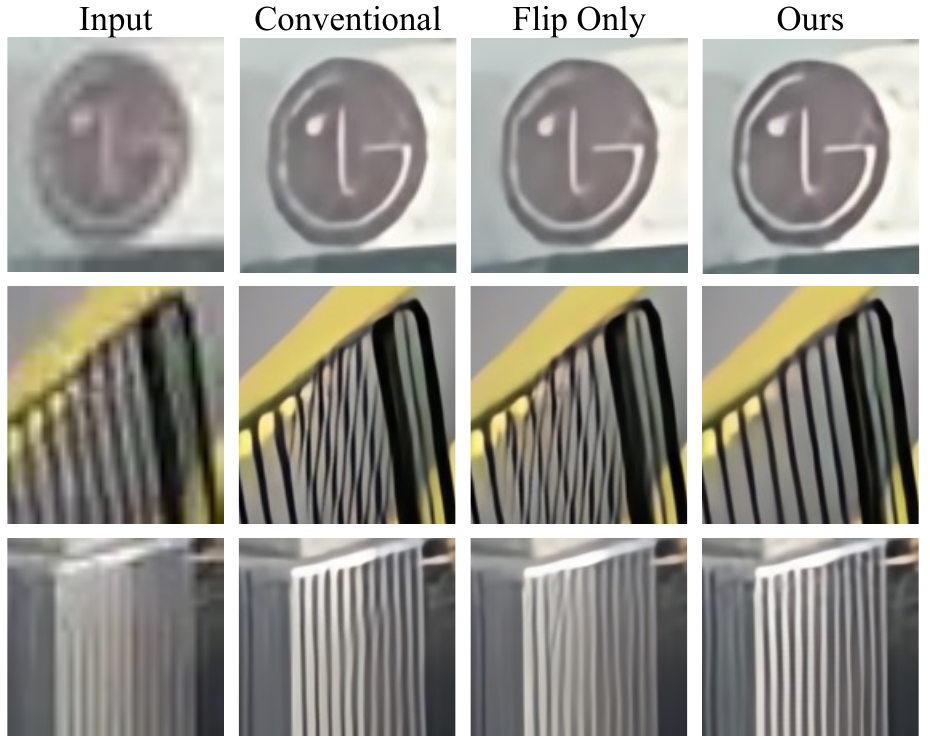}
        \caption{\textbf{Results Using Stochastic Degradation.} While directly flipping the sequence results in degraded performance, applying our stochastic degradation scheme leads to improved performance with up to 40\% reduction of training time.}
        \label{fig:stochastic_result}
    \end{center}
    \vspace{-0.7cm}
\end{figure}
\vspace{0.1cm}
\noindent\textbf{Dynamic Refinement.}
In Fig.~\ref{fig:cleaning} we show an example using our dynamic refinement scheme. On one hand, when applying the cleaning module only once, the noises cannot be completely removed despite more details are shown. On the other hand, it is observed that the outputs are unnaturally flat and details disappear when the cleaning module is applied five times. In contrast, with our dynamic refinement scheme, the cleaning stage is halted automatically to avoid over-smoothing. We see that the outputs contain fewer artifacts while preserving necessary details. We observe that at most three iterations are needed in most scenarios.

We then study the effect of the threshold $\theta$ in Fig.~\ref{fig:cleaning_plot}(a). First, our dynamic refinement scheme leads to a significantly lower NIQE for all thresholds we used. Second, it is observed that different choices of thresholds lead to different levels of details, and hence different NIQE. In Fig.~\ref{fig:cleaning_plot}(b) we compare our scheme with fixed numbers of iterations. Our dynamic refinement scheme determines an image-specific threshold, yielding better performance.
It is noteworthy that one can design a more sophisticated decision process, or manually determine the number of passes to the cleaning module. More elaborative designs of the refinement scheme are left as our future work.

\section{Tradeoff in Training}
In real-world VSR, networks are required to deal with diverse degradations, and hence they are usually trained with multiple degradations. As a result, these networks are often trained with increased batch size to produce a stable gradient. Therefore, training real-world VSR networks often require more computational resources than the non-blind counterparts.
In this work, we delve into two challenges induced by the increased computational budgets, namely 1) speed-performance tradeoff and 2) batch-length tradeoff.

\subsection{Training Speed vs. Performance}
When training with batch size $B$ and sequence length $L$, the CPU needs to load $B{\times}L$ images in each iteration. With increased $B$ in real-world VSR, severe I/O bottleneck is introduced, substantially slowing down training.
Usually, the bottleneck is circumvented by reducing either the batch size or sequence length, resulting in degraded performance. In this work, we propose a stochastic degradation scheme, which significantly improves the training speed without sacrificing performance. The graphical illustration of $L{=}4$ is shown in Fig.~\ref{fig:stochastic}.

In our stochastic degradation scheme, instead of loading $L$ frames in each iteration, we load $L{/}2$ frames and flip the sequence temporally. This design allows us to train with sequences with the same length while reducing the workload of the CPU by half. However, in such a setting, the network perceives content with less variation, and the network can potentially make use of the shortcut that the sequences are temporally flipped.
To improve the diversity of data, we model the degradations to each frame as a \textit{random walk}. Specifically, let $p_i$ be the parameters corresponding to the degradations applied to the $i$-th frame, we have
\begin{equation}
    \label{eq:stoc}
    p_{i+1} = p_i + r_{i+1}.
\end{equation}
Here $r_{i+1}$ represents the differences between the parameters for the $(i{+}1)$-th and $i$-th frames.

As shown in Fig.~\ref{fig:stochastic_result}, when compared to the conventional training scheme, directly flipping the sequence results in similar or degraded performance qualitatively. For instance, the orientations of the line patterns are distorted due to the aliasing effect in the inputs. When our stochastic degradation scheme is applied, the network is more robust to the variation of degradations, leading to improved performance. In addition, as depicted in Table~\ref{tab:stochastic}, by reducing the number of images processed, the workload of the CPUs is significantly reduced. As a result, the I/O bottleneck is ameliorated, and the training time is reduced by up to 40\%\footnote{Different hardware could lead to different levels of bottleneck, and hence different levels of speedup.} without sacrificing performance.

\begin{table}[t]
    \caption{\textbf{Comparison to Stochastic Degradation Scheme.} Our scheme leads to 40\% reduction of training time while maintaining comparable performance.}
    \label{tab:stochastic}
    \begin{center}
        \tabcolsep=0.15cm
        \vspace{-0.5cm}
        \scalebox{0.85}{
            \begin{tabular}{l|c|c}
                                       & Time per iteration $\downarrow$ & NIQE $\downarrow$ \\\hline
                Conventional Scheme    & $\sim$2.5s                      & 4.7191            \\
                Flip Only              & $\sim$1.5s                      & 4.6926            \\
                Stochastic Degradation & $\sim$1.5s                      & 4.6836            \\
            \end{tabular}}
        \vspace{-0.4cm}
    \end{center}
\end{table}

\begin{figure}[t]
    \begin{center}
        \includegraphics[width=0.45\textwidth]{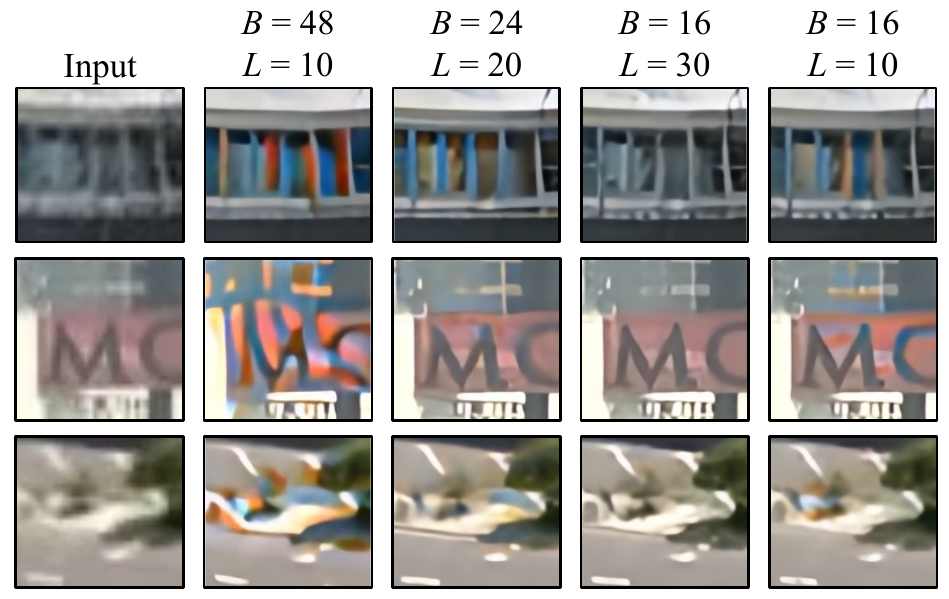}
        \caption{\textbf{Tradeoff Between Batch and Length.} With a fixed computational constraint, training with large batch size and short sequence results in color artifacts and blurrier outputs. Surprisingly, when the length is small, training with large batch size harms the performance.}
        \label{fig:batchlength}
    \end{center}
    \vspace{-0.6cm}
\end{figure}
\begin{table*}[t]
    \caption{\textbf{Quantitative Comparison.} RealBasicVSR obtains the best performance on all four metrics than existing methods with faster speed. Runtime is computed with an output size of $720{\times}1280$, with an Nvidia V100 GPU. \gf{Green} and \bd{blue} colors represent the best and second best performance, respectively.}
    \label{tab:quan}
    \begin{center}
        \tabcolsep=0.15cm
        \vspace{-0.5cm}
        \scalebox{0.75}{
            \begin{tabular}{l|c|c||c|c|c|c|c|c|c}
                                                                              & Bicubic & BasicVSR++~\cite{chan2021basicvsrpp} & RealVSR~\cite{yang2021real} & DAN~\cite{luo2020unfolding} & DBVSR~\cite{pan2021deep} & BSRGAN~\cite{zhang2021designing} & Real-ESRGAN~\cite{wang2021real} & RealSR~\cite{ji2020real} & \textbf{RealBasicVSR} \\\hline
                Params (M)                                                    & -       & 7.3                                  & \gf{2.7}                    & \bd{4.3}                    & 25.5                     & 16.7                             & 16.7                            & 16.7                     & 6.3                   \\
                Runtime (ms)                                                  & -       & \bd{77}                              & 1082                        & 185                         & 239                      & 149                              & 149                             & 149                      & \gf{63}               \\\hline
                NRQM~\cite{ma2017learning} {\color{Green}$\uparrow$}          & 2.8545  & 3.6807                               & 2.5322                      & 3.4347                      & 3.4850                   & \bd{5.8197}                      & 5.8129                          & 5.7030                   & \gf{6.1408}           \\
                NIQE~\cite{mittal2013making} {\color{BrickRed}$\downarrow$}   & 5.2762  & 4.3424                               & 4.9484                      & 4.7844                      & 4.5383                   & 3.2216                           & 3.1263                          & \bd{3.0285}              & \gf{2.5693}           \\
                PI~\cite{blau2018pirm} {\color{BrickRed}$\downarrow$}         & 6.2109  & 5.3309                               & 6.2081                      & 5.6749                      & 5.5267                   & 3.7010                           & \bd{3.6567}                     & 3.6628                   & \gf{3.2143}           \\
                BRISQUE~\cite{mittal2011blind} {\color{BrickRed}$\downarrow$} & 55.225  & 50.665                               & 55.317                      & 51.875                      & 50.937                   & \bd{27.832}                      & 30.679                          & 29.638                   & \gf{27.697}           \\
            \end{tabular}}
        \vspace{-0.7cm}
    \end{center}
\end{table*}

\subsection{Batch Size vs. Sequence Length}
With a fixed computation budget, the increased batch size when training real-world VSR models inevitably leads to a decrease in sequence length. On one hand, training with a larger batch size enables the network to perceive more degradations and scene content in each iteration, leading to more stable gradients. On the other hand, training with longer sequences allows the network to employ long-term information more effectively.
However, one must choose between a larger batch or a longer sequence when computational resources are limited. We are interested in the tradeoff between them, with an aim to provide an effective setting for future works. In this section, we train RealBasicVSR with a constraint of $B{\times}L{=}480$ and discuss the performance of these models. Our stochastic degradation scheme is used.

As shown in Fig.~\ref{fig:batchlength}, when training with $B{=}48$, $L{=}10$, it is observed that the outputs contain severe color artifacts and distorted details. This undesirable effect reduces when we increase the sequence length. In particular, the color artifacts are significantly reduced when $L$ increases from $10$ to $20$, and are further eliminated when $L$ increases to $30$.

The above comparison shows that training with longer sequences is preferable. We speculate that networks trained with short sequences cannot adapt to long sequences during inference, due to the domain gap between training and inference.
To further demonstrate that the importance of long sequences in training, we fix $B$ to $16$ and reduce $L$ from $30$ to $10$. It is observed that the corresponding regions shows the same color artifacts and blur when $L$ is reduced. Therefore, it is suggested to employ a longer sequence when a computational constraint is imposed.

\begin{figure}[t]
    \begin{center}
        \includegraphics[width=0.44\textwidth]{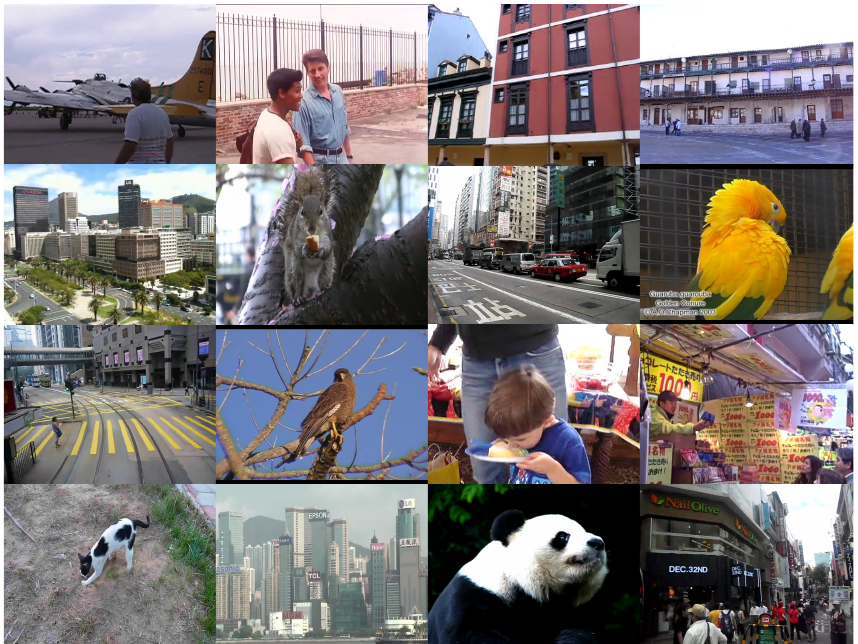}
        \caption{\textbf{VideoLQ Dataset.} Our \textit{VideoLQ} dataset consists of videos with a wide range of content and resolutions, collected from different video hosting sites such as Flickr and YouTube. It can be served as a common ground for future comparison.}
        \label{fig:videolq}
    \end{center}
    \vspace{-0.85cm}
\end{figure}
\begin{figure*}[t]
    \begin{center}
        \includegraphics[width=0.99\textwidth]{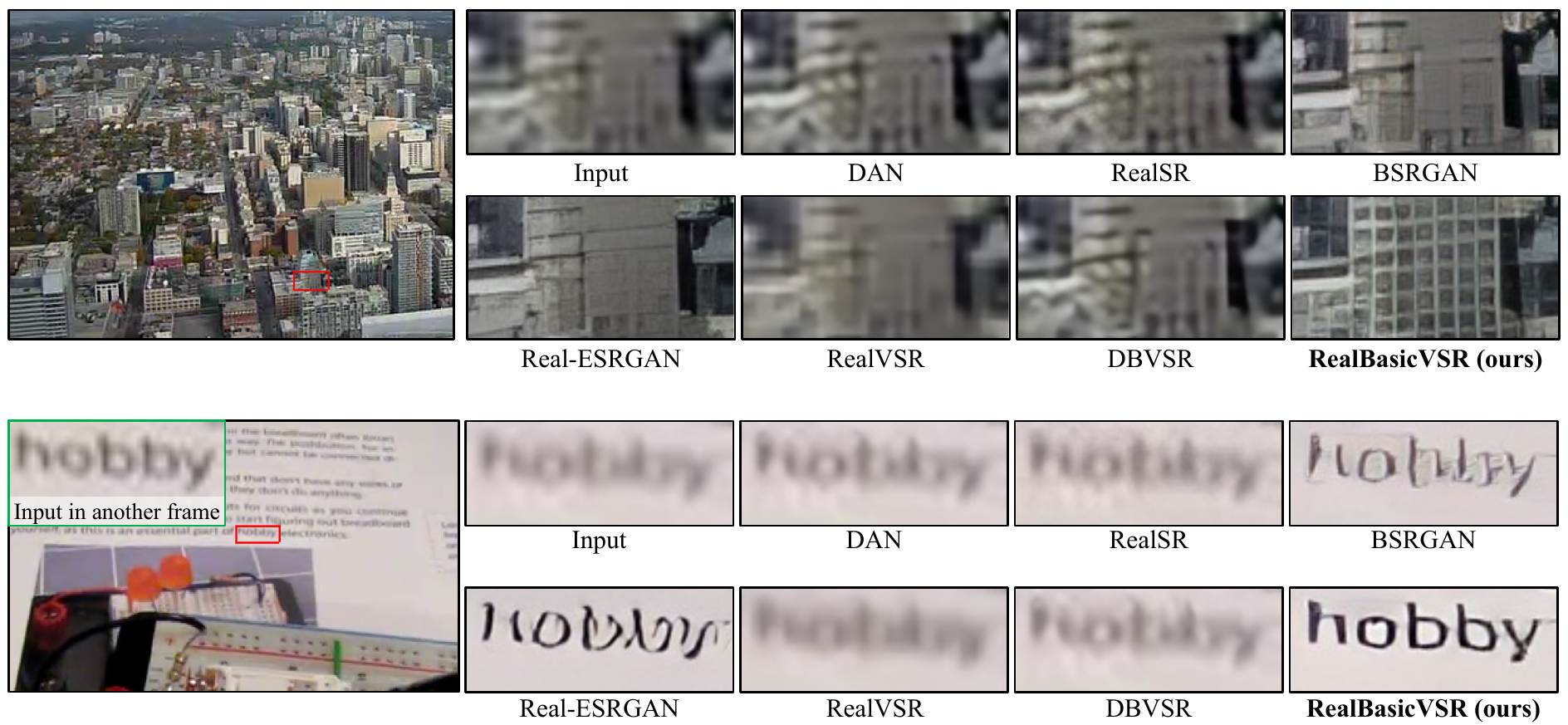}
        \caption{\textbf{Qualitative Comparison.} Our RealBasicVSR is able to aggregate long-term information effectively. It generates much more details when compared to existing works. In particular, by aggregating long-term information through propagation, RealBasicVSR successfully restores the word ``hobby'', which can be clearly seen in latter frames . \textbf{(Zoom-in for best view)}}
        \label{fig:quali}
    \end{center}
    \vspace{-0.7cm}
\end{figure*}

\section{VideoLQ Dataset and Benchmark}
To assess the generalizability of real-world VSR methods, a benchmark that covers a wide range of \textit{degradations}, \textit{content}, and \textit{resolution} is indispensable. Most existing datasets~\cite{liu2014bayesian,nah2019ntire,xue2019video,yi2019progressive} focus only on synthetic degradations such as bicubic downsampling, and hence they cannot reflect the efficacy of real-world VSR methods.
The recent RealVSR dataset~\cite{yang2021real} consists of LR-HR pairs of videos captured by the dual-camera system in iPhone. Although the data is not constructed by synthetic degradations, the sequences are captured by a single camera, and hence the LR videos contain only camera-specific degradations. Hence, there is no guarantee that methods performing superiorly in the RealVSR dataset can generalize to videos in the wild.

In this work, we propose the \textit{VideoLQ} dataset. Examples of the videos are shown in Fig.~\ref{fig:videolq}.
The videos in our VideoLQ dataset are downloaded from various video-hosting sites such as Flickr and YouTube, with a Creative Common license. To ensure diversity of the videos, we select videos with different resolutions and contents to cover as many degradations as possible.
For each video, we extract a sequence of 100 frames with no scene changes allowed, so that methods relying on long-term propagation can be assessed. The sequences are selected to contain enough textures or texts for ease of comparison. Additionally, the ground-truth videos in Vid4~\cite{liu2014bayesian} are also included.


\subsection{Experimental Settings}
We conduct experiments on our VideoLQ dataset. We compare our RealBasicVSR with seven state of the arts, including four image models: RealSR~\cite{ji2020real}, DAN~\cite{luo2020unfolding}, Real-ESRGAN~\cite{wang2021real}, BSRGAN~\cite{zhang2021designing} and three video models: BasicVSR++\footnote{Trained with bicubic downsampling as a reference.}~\cite{chan2021basicvsrpp}, RealVSR~\cite{yang2021real}, DBVSR~\cite{pan2021deep}. More discussion are provided in the supplementary material.

\noindent\textbf{Training Degradations.}
Following Real-ESRGAN~\cite{wang2021real}, we adopt the second-order order degradation model, and we apply random blur, resize, noise, and JPEG compression as image-based degradations. In addition, we incorporate video compression, which is a common technique to reduce video size. Unlike the aforementioned degradations, video compression implicitly considers the inter-dependencies between video frames, providing us with temporally and spatially varying degradations. We apply compression with randomly selected codecs and bitrates during training, and we observe performance gain with video compression included. The detailed settings are provided in the supplementary material. For the methods in comparison, we use their publicly available code.

\noindent\textbf{Training Settings.}
Following DBVSR~\cite{pan2021deep}, we use the REDS dataset~\cite{nah2019ntire} for training.
We adopt Adam optimizer~\cite{kingma2014adam} with constant learning rates. The patch size of input LR frames is $64{\times}64$. We apply our stochastic degradation scheme with temporal length 30\footnote{That means, the CPU loads $15$ images in each iteration.}. The training is divided into two stages:
We first pre-train RealBasicVSR with only output loss and image cleaning loss for 300K iterations, with batch size 16 and learning rate $10^{-4}$.
We then finetune the network with also perceptual loss~\cite{johnson2016perceptual} and adversarial loss~\cite{goodfellow2014generative} for 150K iterations. The batch size is reduced to $8$. The learning rates of the generator and discriminator are set to $5{\times}10^{-5}$ and $10^{-4}$.

\noindent\textbf{Architecture.}
In the adversarial training, we use RealBasicVSR as the generator, and adopt the discriminator of Real-ESRGAN.
For the generator, our image cleaning module $C$ consists of 20 residual blocks. We use BasicVSR as our VSR network $S$, with the number of residual blocks set to 40. The number of feature channels is $64$. Detailed experimental settings and model architectures are provided in the supplementary material.

\noindent\textbf{Quantitative Metric.}
As ground-truths are not available for real-world videos, common metrics such as PSNR and SSIM cannot be used in this task. Therefore, we adopt four commonly used non-reference metrics NIQE~\cite{mittal2013making}, NRQM~\cite{ma2017learning}, PI~\cite{blau2018pirm}, and BRISQUE~\cite{mittal2011blind} to supplement our qualitative comparison.

\subsection{Comparison to State of the Arts}
We show real-world examples on our VideoLQ dataset in Fig.~\ref{fig:quali}. Equipped with the image cleaning module, RealBasicVSR is able to aggregate long-term information through propagation effectively. As a result, it generates much more details in fine regions, improving visual quality. For instance, only RealBasicVSR is able to recover the word ``hobby'', which can be clearly seen in other frames. 

In addition to qualitative results, we also provide quantitative measures as a reference. When compared to existing methods, RealBasicVSR achieves better performance on all metrics with faster speed. In particular, RealBasicVSR outperforms the recent RealVSR~\cite{yang2021real} with 17${\times}$ faster speed. When compared to Real-ESRGAN~\cite{wang2021real}, which uses a similar training pipeline, RealBasicVSR performs superiorly with lower complexity and faster speed.

The above methods employ only either single-image or short-term information. While these methods demonstrate significant improvements in terms of degradation removal, they cannot effectively recover the details beyond the input image and its local neighbors, which require aggregation of information from distant frames. In contrast, RealBasicVSR explores the possibility of exploiting long-term information in real-world VSR, and both our qualitative and quantitative results show the effectiveness of RealBasicVSR in exploiting such information for detail synthesis.

\section{Discussion}

A common belief in existing VSR studies~\cite{chan2021basicvsr,chan2021basicvsrpp} is that long-term propagation is beneficial to restoration performance. Yet, such discussion is limited to non-blind VSR.
In this work, we examine the contributions of temporal propagation in real-world VSR and find that long-term information is also beneficial to this task but do not come for free, due to the diverse and complicated degradations in the wild.
As an explorational study, we reveal several challenges in real-world VSR. We find that the domain gap on degradations and the increased computational costs result in various challenges and tradeoffs. We then provide respective solutions to the challenges including the cleaning module and stochastic degradation scheme, which are easy to implement.
We hope our study and findings in our work as well as our VideoLQ dataset will lay a good foundation and inspire future works in real-world VSR.

{\small
  \bibliographystyle{ieee_fullname}
  \bibliography{short,bib.bib}
}
\clearpage
\appendix
\section{Architecture and Experimental Settings}
\noindent\textbf{Architecture.}
We use a simple architecture in this work for explorational purpose. First, a convolution is used to extract shallow features from the input image. A stack of 20 residual blocks are then used to extract deep features. A final convolutional layer is then used to produce the clean image. We adopt BasicVSR~\cite{chan2021basicvsr} as the VSR network. We reduce the number of residual blocks from 60 to 40 to maintain comparable complexity to the original BasicVSR.
\vspace{0.15cm}

\noindent\textbf{Loss Function.}
For the output fidelity loss $\mathcal{L}_{pix}$ and image cleaining loss $\mathcal{L}_{clean}$, we use Charbonnier loss~\cite{charbonnier1994two} since it better handles outliers and improves the performance over the conventional $\ell_2$ loss~\cite{lai2017deep}.
In addition, we use perceptual loss~\cite{johnson2016perceptual} $\mathcal{L}_{per}$ and adversarial loss~\cite{goodfellow2014generative} $\mathcal{L}_{adv}$ to achieve better visual quality.

In the first stage, we pretrain the generator (\ie, RealBasicVSR) with the fidelity loss and image cleaning loss:
\begin{equation}
    \mathcal{L}_{1st} = \mathcal{L}_{pix} + \mathcal{L}_{clean}.
\end{equation}
We then finetune the network with also perceptual loss and adversarial loss:
\begin{equation}
    \mathcal{L}_{2nd} = \mathcal{L}_{pix} + \mathcal{L}_{clean} + \lambda_{per}\mathcal{L}_{per} + \lambda_{adv}\mathcal{L}_{adv}.
\end{equation}
In our experiments, $\lambda_{per}{=}1$ and $\lambda_{adv}{=}5{\times}10^{-2}$. Note that in the second stage, the weights of the cleaning module are kept fixed.

\vspace{0.15cm}
\noindent\textbf{Training Degradations.}
Following Real-ESRGAN~\cite{wang2021real}, we adopt the second-order order degradation model, and we apply random blur, resize, noise, and JPEG compression as image-based degradations. In addition, we incorporate video compression, which is a common technique to reduce video size. Unlike the aforementioned degradations, video compression implicitly considers the inter-dependencies between video frames, providing us with temporally and spatially varying degradations.
The settings of image-based degradations follow Real-ESRGAN~\cite{wang2021real}. For the video compression, in each iteration, we randomly select one of the following codecs: ``libx264'', ``h264'', and ``mpeg4''. The bitrate is uniformly selected from the range $[10^{4}, 10^5]$. Video compression is added right after JPEG compression.

\vspace{0.15cm}
\noindent\textbf{Implementation.}
We implement our models with PyTorch and train the models using eight NVIDIA Tesla V100 GPUs. Code will be made publicly available.

\section{Discussion of Baselines}
In this work, we compare our RealBasicVSR with seven state of the arts, including four image models: RealSR~\cite{ji2020real}, DAN~\cite{luo2020unfolding}, Real-ESRGAN~\cite{wang2021real}, BSRGAN~\cite{zhang2021designing} and three video models: BasicVSR++\footnote{Trained with bicubic downsampling, as a reference.}\cite{chan2021basicvsrpp}, RealVSR~\cite{yang2021real}, DBVSR~\cite{pan2021deep}. They are representative methods in image and video super-resolution that achieve promising performance.

With specific designs in training, these methods demonstrate significant improvements when compared to non-blind methods. However, while these methods succeed in removing degradations in the input images, they are inferior in recovering details beyond the image itself or its local neighbors, due to the fact that they do not exploit long-term information available in videos.

\begin{figure*}[t]
    \begin{center}
        \includegraphics[width=0.99\textwidth]{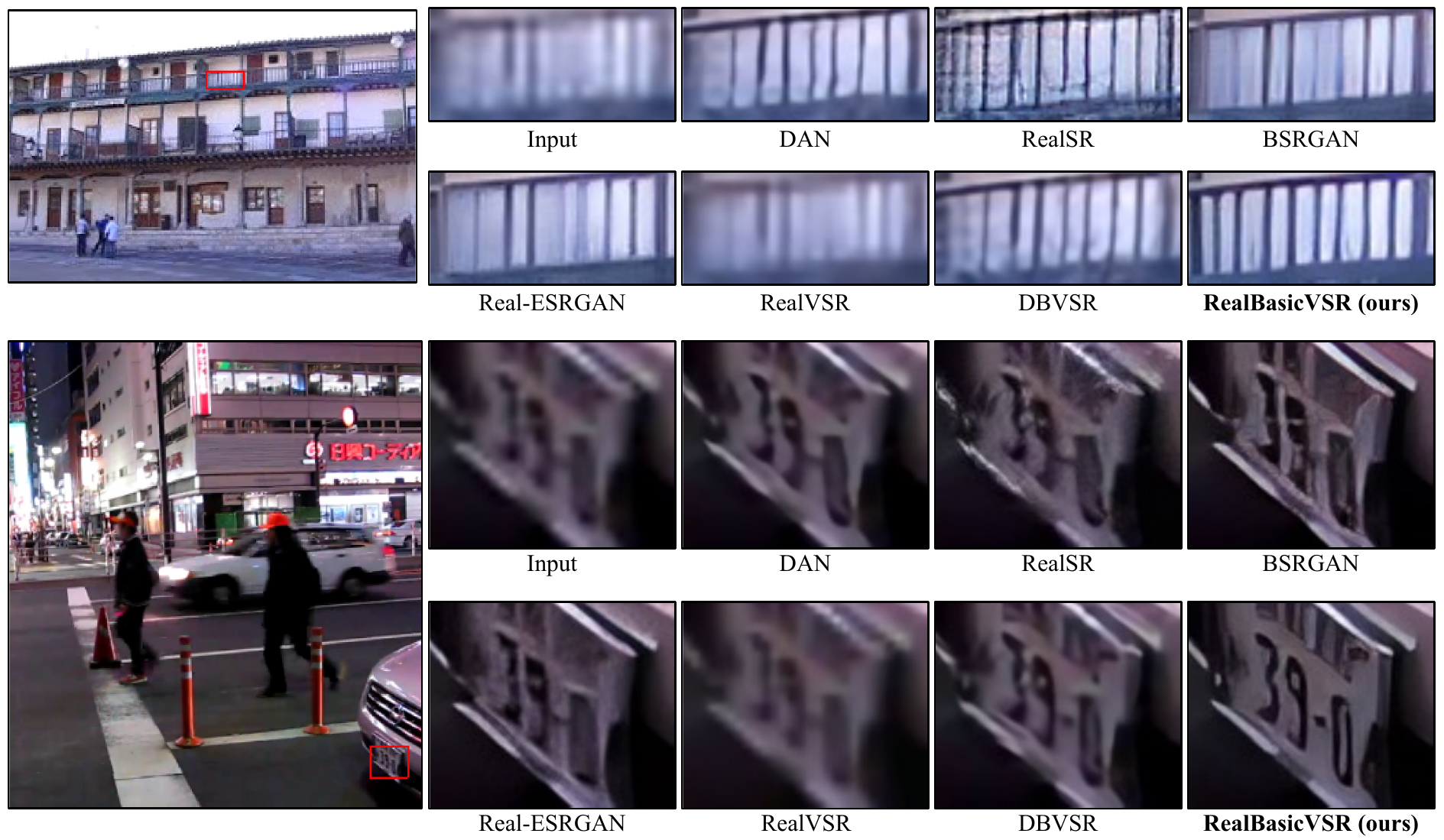}
        \caption{\textbf{Qualitative Comparison.} By employing the long-term information effectively, RealBasicVSR restores more details when compared to existing state of the arts. \textbf{(Zoom-in for best view)}}
        \label{fig:supp_quali_1}
    \end{center}
    \vspace{-0.5cm}
\end{figure*}

\begin{figure*}[t]
    \begin{center}
        \includegraphics[width=0.99\textwidth]{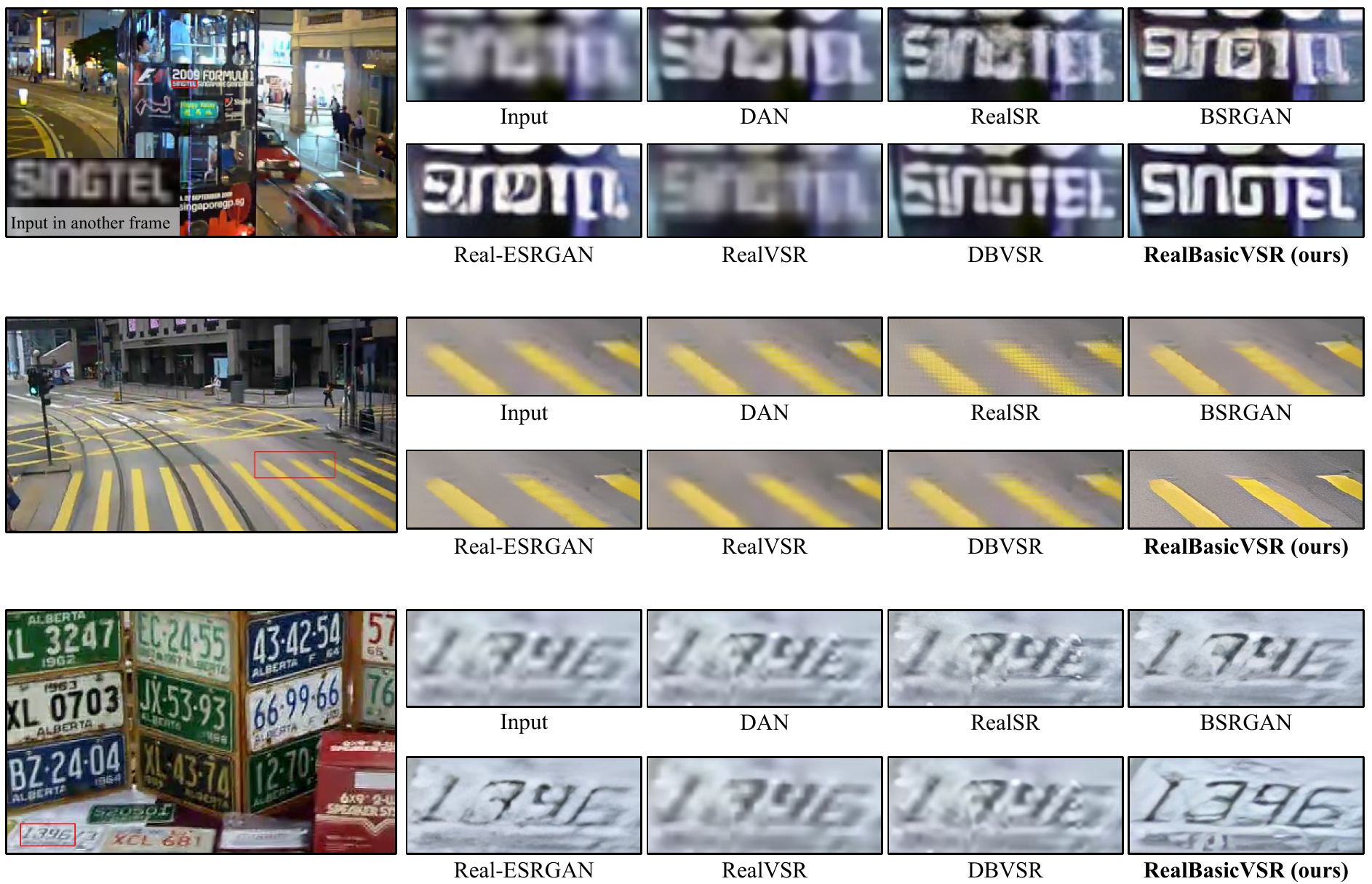}
        \caption{\textbf{Qualitative Comparison.} By employing the long-term information effectively, RealBasicVSR restores more details when compared to existing state of the arts. \textbf{(Zoom-in for best view)}}
        \label{fig:supp_quali_2}
    \end{center}
    \vspace{-0.5cm}
\end{figure*}

Despite being extensively discussed in non-blind VSR, the use of long-term information has not been explored in real-world VSR. In this work, we find that such long-term information, if used with designated designs, is also useful in real-world VSR. With the benefits of our findings and designs, RealBasicVSR is able to restore more details than the methods in comparison, as shown in Fig.~\ref{fig:supp_quali_1} and Fig.~\ref{fig:supp_quali_2}.

\section{Dynamic Refinement}
\begin{figure*}[t]
    \begin{center}
        \includegraphics[width=0.99\textwidth]{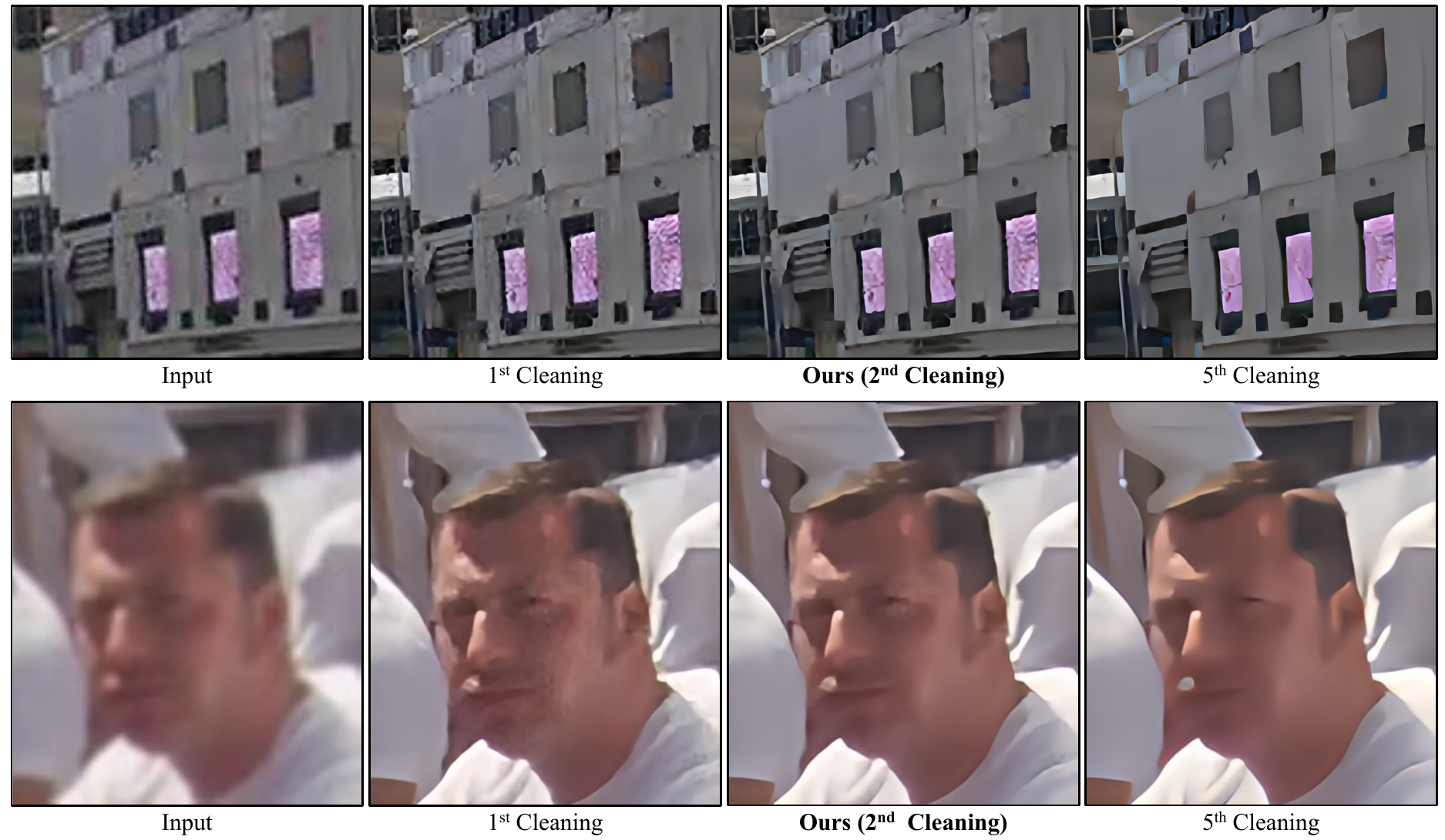}
        \caption{\textbf{Dynamic Refinement.} Our dynamic refinement scheme removes remaining noises and artifacts in the first cleaning while avoiding over-smoothing. \textbf{(Zoom-in for best view)}}
        \label{fig:supp_refinement}
    \end{center}
    \vspace{-0.5cm}
\end{figure*}

In this section, we show additional examples demonstrating the effects of our dynamic refinement. As shown in Fig.~\ref{fig:supp_refinement}, unpleasant artifacts remain in the outputs when applying cleaning once, and unnatually flat outputs due to over-cleaning are observed when our cleaning module is applied five times. In contrast, our refinement scheme automatically stops the refinement to avoid over-smoothing while cleaning excessive artifacts, leading to improved performance. More sophisticated decision processes are left as our future work.


\end{document}